\documentclass[journal]{IEEEtran}

\ifCLASSINFOpdf
\else
   \usepackage[dvips]{graphicx}
\fi
\usepackage{url}

\usepackage{graphicx}
\usepackage{subfig}
\usepackage{amssymb}
\usepackage{bbding}
\usepackage{amsmath}
\usepackage{amsmath}
\usepackage{booktabs}
\usepackage{multirow}
\usepackage{cases}
\usepackage[colorlinks,
linkcolor=red,
anchorcolor=blue,
citecolor=green,
]{hyperref}

\usepackage[numbers,sort&compress]{natbib}

\newsavebox{\measurebox}

\begin{document}

\title{Residual Swin Transformer Channel Attention Network for Image Demosaicing}

\author{Wenzhu Xing, Karen Egiazarian, \IEEEmembership{Fellow, IEEE}
\thanks{This paper is submitted for review on 13rd of April in 2022.}
\thanks{W. Xing and K. Egiazarian are with the Computational Image Group, Tampere University, Korkeakoulunkatu 7, Finland (e-mail: wenzhu.xing@tuni.fi; karen.eguiazarian@tuni.fi).}}

%\markboth{Journal of \LaTeX\ Class Files, Vol. 14, No. 8, August 2015}
\maketitle

\begin{abstract}
%Image demosaicing plays an essential role in image signal processing pipeline. 
Image demosaicing 
%or color filter array interpolation
is problem of interpolating  full-resolution color images from raw sensor (color filter array)  data.  %color-incomplete image to the full-color image. 
%In recent years, 
During last decade, deep neural networks have been  widely used in image restoration, and in particular, in demosaicing, attaining significant  performance improvement. 
%Thanks to the deep convolutional neural networks, 
%the performance of image demosaicing methods attain unprecedented improvement.
In recent years, vision transformers have been designed and successfully used in  various 
computer vision applications.
%more and more vision transformers   
%are proposed to solve the high-level vision problems and even the 
%image restoration, %tasks
% for example, the Swin Transformer.
%However, \textcolor{red}{they have not yet found successful application in image demosaicing.} 
%the application of the vision transformer in image demosaicing is still lack.
One of the recent methods of image restoration based on a Swin Transformer (ST),  SwinIR, demonstrates  state-of-the-art performance with a smaller number of parameters than neural network-based methods.
Inspired by the success of SwinIR, we propose in this paper a novel Swin Transformer-based network for image demosaicing, called RSTCANet. %based on Swin Transformer.
To extract image features,
%In order to all-round extract features,
RSTCANet stacks several residual Swin Transformer Channel Attention blocks (RSTCAB), introducing
%in where we introduce 
the channel attention for each two successive ST blocks.
Extensive experiments %\textcolor{red}{on image demosaicing}
demonstrate that RSTCANet  %\textcolor{red}{not only} 
outperforms %not only 
state-of-the-art 
%methods on 
image demosaicing methods, and has a smaller number of parameters.
\end{abstract}


\IEEEpeerreviewmaketitle

\section{Introduction and Related Works}\label{sec:rw}
\begin{figure}[!htb]
\centering
\sbox{\measurebox}{
  \begin{minipage}[b]{.21\textwidth}
  \subfloat[Urban100:img026.png]{\includegraphics[width=\textwidth]{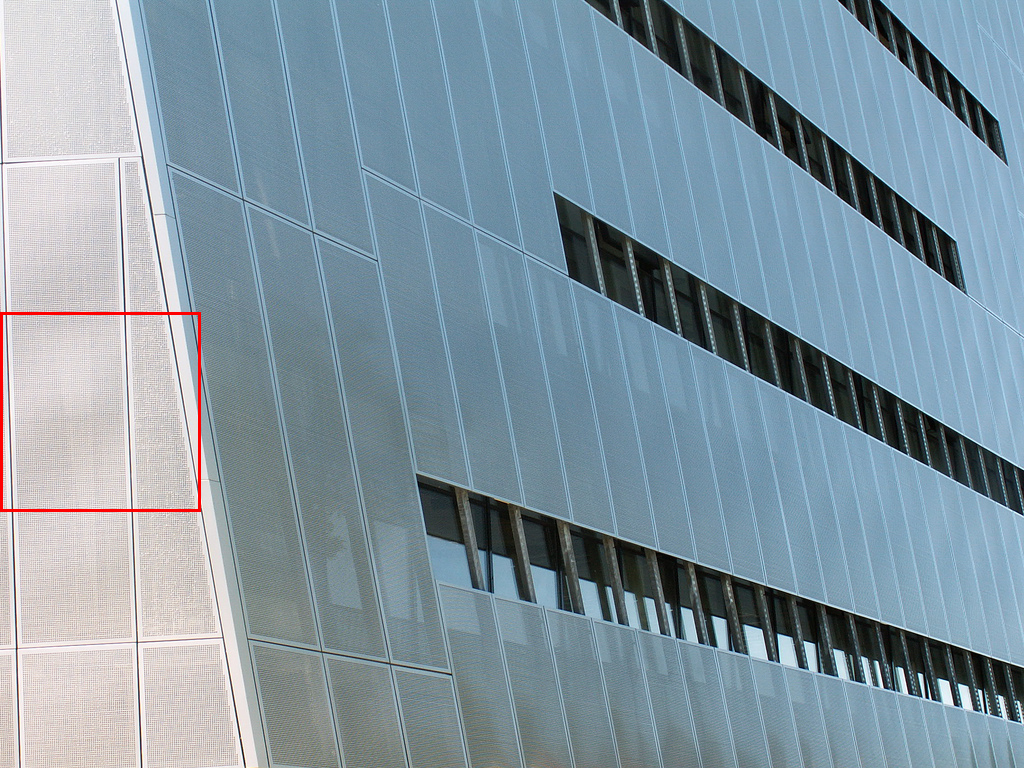}}
  \end{minipage}}
\usebox{\measurebox}\enspace
\begin{minipage}[b]{.058\textwidth}
\centering
\subfloat[]{\includegraphics[width=\textwidth]{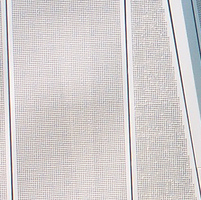}}

\subfloat[]{\includegraphics[width=\textwidth]{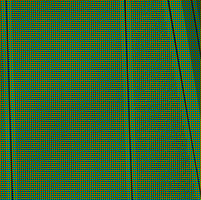}}
\end{minipage}
\begin{minipage}[b]{.058\textwidth}
\centering
\subfloat[]{\includegraphics[width=\textwidth]{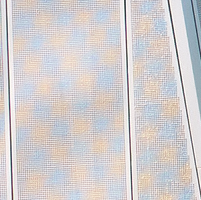}}
%\vfill

\subfloat[]{\includegraphics[width=\textwidth]{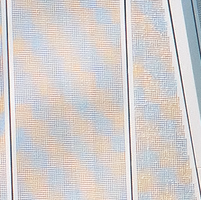}}
\end{minipage}
\begin{minipage}[b]{.058\textwidth}
\centering
\subfloat[]{\includegraphics[width=\textwidth]{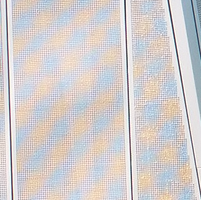}}

\subfloat[]{\includegraphics[width=\textwidth]{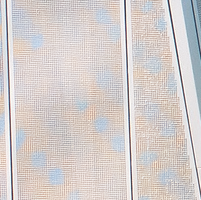}}
\end{minipage}
\begin{minipage}[b]{.058\textwidth}
\centering
\subfloat[]{\includegraphics[width=\textwidth]{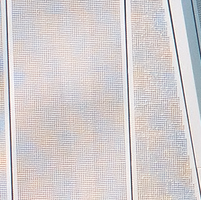}}

\subfloat[]{\includegraphics[width=\textwidth]{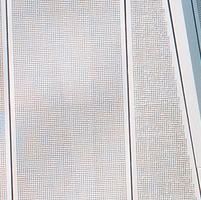}}
\end{minipage}
\caption{Visual results comparison of different demosaicing methods on image \textit{026} from Urban100 dataset. (a)  Ground-truth and selected area; (b) Ground-truth; (c) Mosaiced; (d) IRCNN; (e) RSTCANet-B; (f) DRUNet; (g) RSTCANet-S; (h) RNAN; (i) RSTCANet-L.}
\label{fig:dm_urban_compare}
\vspace{-0.3cm}
\end{figure}

Most modern digital cameras record only one color channel (red, green, or blue) per pixel. 
In order to recover the missing pixels, the image demosaicing models are proposed to reconstruct a full color image from a one-channel mosaiced image.
Existing demosaicing methods can be 
%roughly 
classified into two categories: model-based methods~\cite{hirakawa2005adaptive,malvar2004high,su2006highly,zhang2005color}, 
which recover images based on  mathematical models and image priors in the spatial-spectral domain; 
and learning-based methods~\cite{he2012self,sun2012separable,go2000interpolation,kapah2000demosaicking,syu2018learning,gharbi2016deep,zhang2021plug,zhang2017learning,zhang2019residual}, based on process mapping learned from abundant ground-truth image and mosaiced image pairs.
Among these methods, the recent ones~\cite{zhang2021plug,zhang2017learning,zhang2019residual} attain state-of-the-art performance.
However, there are still color artifacts in their resulting images (Fig.~\ref{fig:dm_urban_compare}), especially in high frequency regions. 
Besides, the cost of these networks to improve performance is to increase the depth of the network, which results in a bigger model size (Table~\ref{tab:compare}).

To overcome the above-mentioned problems,
%solve the problems mentioned above,
we turn our attention to other lighter but effective models for image restoration.
Recently proposed
%, a new 
vision transformer, called Swin Transformer~\cite{liu2021swin} %, is proposed and 
 outperforms 
state-of-the-art
%other methods 
in several vision problems,
such as image classification, object detection, and semantic segmentation. Same
%In the same 
year, a U-Net method based on Swin Transformer 
%is 
has been
proposed for medical image segmentation, called Swin-Unet~\cite{cao2021swin}.
Meanwhile, another Swin Transformer-based method, SwinIR~\cite{liang2021swinir}, was 
%is 
proposed for image restoration. SwinIR surpasses state-of-the-art methods on image super-resolution, image denoising, and JPEG compression artifact reduction with fewer number of parameters.
Inspired by the success of SwinIR, we adopt Swin Transformer 
to propose a lightweight model for image demosaicing.
We notice that while utilizing Swin Transformer in SwinIR is helpful to fully excavate the image features patch attentions horizontally, %. However, 
an extraction of the channel features  vertically has not received equal attention. 

\begin{figure*}[!ht]
    \centering
    \includegraphics[width=0.8\linewidth]{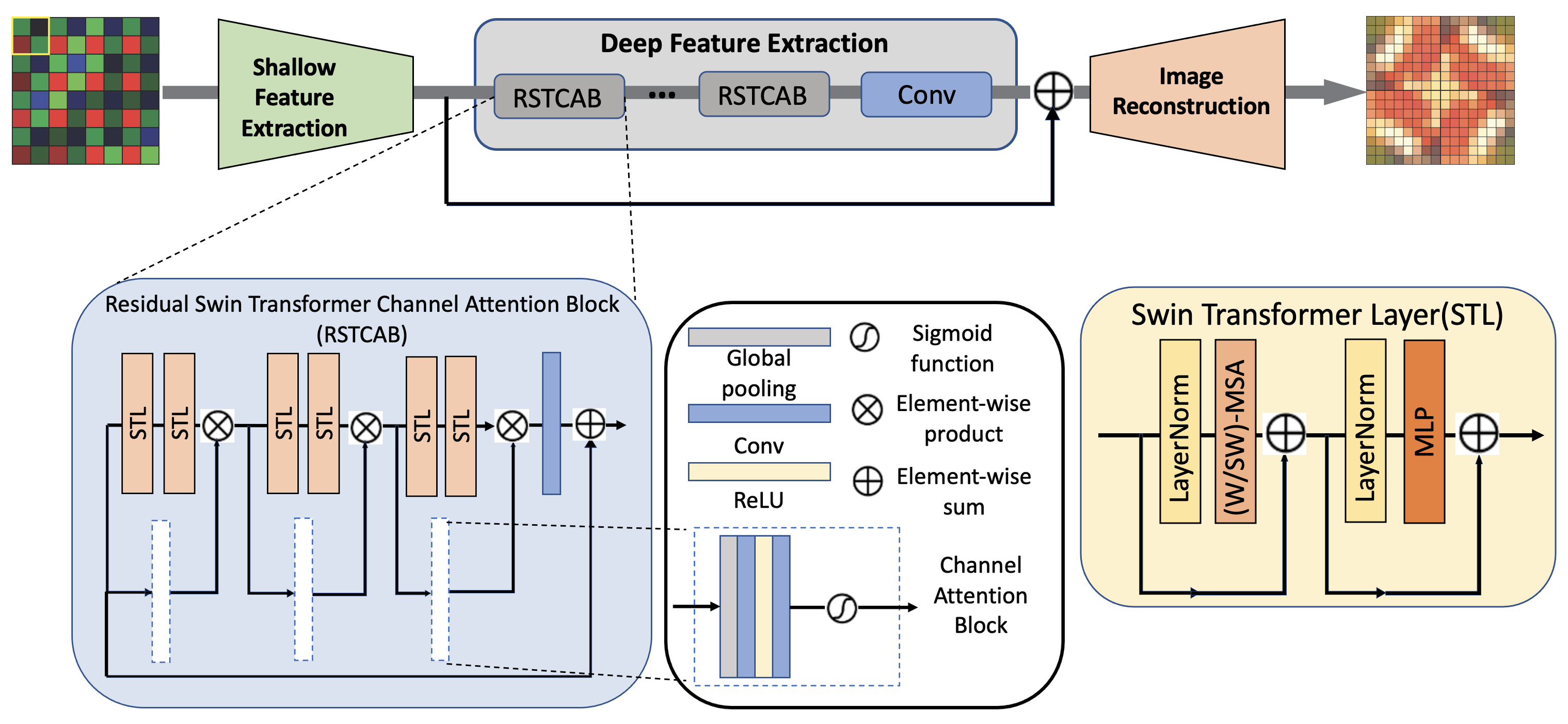}
    \caption{Residual Swin Transformer Channel Attention Network (RSTCANet) and Residual Swin Transformer Channel Attention Block (RSTCAB).}
    \label{fig:RSTCANet}
    \vspace{-0.5cm}
\end{figure*}

Considering the inter-dependencies among the feature channels should be utilized  %noticed 
as well, 
we introduce the channel attention~\cite{zhang2018image} in the basic block of SwinIR, residual Swin Transformer block (RSTB), 
to comprehensively extract image features.
The proposed combination is named RSTCAB, which has six Swin Transformer layers (STL) and three channel attention blocks. 
For each two successive STLs, one channel attention block is utilized to generate different attention for each channel-wise feature learned by STLs.
The channel attention (CA) is first proposed in RCAN~\cite{zhang2018image}.
It consists of a GlobalPooling layer, a down-sampling convolution layer, an  up-sampling convolution layer, and the sigmoid function.
The ablation study in Sec.~\ref{subsec:as} proves the adoption of CA can further improve the performance of RSTB on demosaicing.

There is a recent work on %concurrent work 
SCUNet~\cite{zhang2022practical}, 
%proposed 
a U-Net based on Swin Transformer, for a  blind denoising.
The basic module SC block of SCUNet combines the Swin Transformer and residual convolutional block.
In contrast, our proposed RSTCAB introduces channel attention blocks in the Swin Transformer blocks.
In addition, the application of U-Net architecture in RSTCANet drops the performance on image demosaicing but increases the number of parameters (see Sec.~\ref{subsec:as}).

In summary, there are three main contributions in this paper: 
(1) We propose the first vision transformer-based method RSTCANet for image demosaicing; 
(2) The proposed residual Swin Transformer channel attention block (RSTCAB) takes advantage of both Swin Transformer and channel attention. 
Compared with either other Swin Transformer-based block~\cite{liang2021swinir} or residual channel attention block~\cite{zhang2018image}, RSTCAB attains the best performance on image demosaicing;
(3) RSTCANet achieves state-of-the-art performance on four datasets with smaller model size compared with the existing image demosaicing methods. 
In addition, the resulting images generated by RSTCANet contain much less visible artifacts (see an example in Fig.~\ref{fig:dm_urban_compare}).

%\section{Related Works}\label{sec:rw}
%\input{RSTCANet/tex/related_works}

\section{Method}
\subsection{Framework}

\textbf{Network architecture.} The architecture of our proposed residual Swin Transformer Channel Attention network (RSTCANet) is shown in Fig.~\ref{fig:RSTCANet}.
Similar to SwinIR~\cite{liang2021swinir}, the network consists of three modules: the shallow feature extraction, 
%module, 
the deep feature extraction,
%module 
and the image reconstruction modules.
The shallow feature extraction module is composed of a pixel shuffle layer and a vanilla linear embedding layer. 
For deep feature extraction, we propose residual Swin Transformer Channel Attention blocks (RSTCAB) to extract both hierarchical window based self-attention-aware features~\cite{liu2021swin} and vertical channel-attention-aware features.
This module consists of $K$ RSTCAB and one $3\times3$ convolutional layer.
The shallow and deep features are first aggregated by a long skip connection before they fed into the image reconstruction module. 
The image reconstruction consists of the up-sampling
%simply made up by an up-sample
layer and two $3\times3$ convolutional layers.

\textbf{Loss function.} We optimize the RSTCANet with the $\mathcal{L}_1$ loss function.
Given the training pairs $\{I_M^i, I_{GT}^i\}_{i=1}^N$, %which contain
containing
 $N$ mosaiced inputs and their corresponding ground truth images, the optimization of the parameters of RSTCANet can be formulated as :
\begin{equation}
    \mathcal{L}(\Theta) =  \frac{1}{N}\sum_{i=1}^N\Vert RSTCANet(I_M^i)-I_{GT}^i\Vert_1,
\end{equation}
where $\Theta$ denotes the parameter set of RSTCANet and $\Vert\cdot\Vert_1$ denotes $\ell_1$ norm.

\subsection{Residual Swin Transformer Channel Attention Block}

As shown in Fig.~\ref{fig:RSTCANet}, there are $N$ Swin Transformer layers (STL) and $N/2$ channel attention blocks (CA), and one $3\times3$ convolutional layer in our proposed residual Swin Transformer Channel Attention block (RSTCAB). 
There is also a skip connection in the RSTCAB, 
%as well. 
%It can guarantee 
guaranteeing that the RSTCAB 
%to
will focus on the differences between the input and output images.
The skip connection in RSTCAB and the long skip connection in the network results in the proposed RSTCANet becoming a residual in residual framework, as in many other image restoration methods~\cite{zhang2019residual,zhang2018image,zhang2020residual}.

For each two successive STL, the channel attention block generates the channel statistics with the input of two STLs and multiplies the produced attention with the output of two STLs. 
The $N$ channel attention blocks in the same RSTCAB share parameters.
The structure of the channel attention block is same as the one in \cite{zhang2018image}, %which we have mentioned in 
see Sec.~\ref{sec:rw}). 
With the channel attention blocks, the residual component learned by STL in the RSTCAB is adaptively rescaled. 
The convolutional layer at the end of RSTCAB is very important 
for the vision transformer-based image restoration network.
We prove this in Sec.~\ref{subsec:as}.
The discussions about the effect and the number of the channel attention blocks are also included in Sec.~\ref{subsec:as}.

\subsection{Architecture Variants}\label{subsec:av}

In order to fairly compare our proposed RSTCANet with %other 
state-of-the-art image demosaicing methods, we build three different model variants,  like in~\cite{liu2021swin}.
We build our base model, called RSTCANet-B, to have the smallest model size and computation complexity. 
We also introduce RSTCANet-S and RSTCANet-L, which are versions of about $3\times$ and $6\times$ the model size and computational complexity, respectively.
The 
%detailed 
parameter settings of these model variants are shown in Table~\ref{tab:mv}.
%listed as following:
\begin{table}[!ht]
    \centering
    \caption{The parameter settings of different model variants. C is the channel number. $K$ and $N$ denote the number of RSTCAB and the number of STL in one RSTCAB, respectively.}
    \label{tab:mv}
    \resizebox{0.8\linewidth}{!}{
    \begin{tabular}{l|c c c c c}
    \toprule
    Model Variants & C & $K$ & Multihead & $N$ \\\hline
    \hline
    RSTCANet-B & 72 & 2 & 6 & 6 \\\hline
    RSTCANet-S & 96 & 4 & 6 & 6 \\\hline
    RSTCANet-L & 128 & 4 & 8 & 8 \\
    \bottomrule
    \end{tabular}}
\end{table}
% \begin{itemize}
%     \item RSTCANet-B: C=72, RSTCAB number: 2, Multihead number: 6
%     \item RSTCANet-S: C=96, RSTCAB number: 4, Multihead number: 6
%     \item RSTCANet-L: C=128, RSTCAB number: 4, Multihead number: 8
% \end{itemize}
The model size of the model variants for image demosaicing is shown in the last column of  Table~\ref{tab:compare}.

\section{Experiments}
%\subsection{Experimental Setup}

For the training, we have applied Nvidia Tesla V100 GPU with 32 GB memory from the Tampere University TCSC Narvi computing cluster. 
We select DIV2K~\cite{Agustsson_2017_CVPR_Workshops} as our training set, which contains 800 training images.
Data augmentation is performed on images, which are randomly rotated by $90^{\circ}$, $180^{\circ}$, $270^{\circ}$ and flipped horizontally.
The batch size is $16$, and the patch size is $64 \times 64$.
For  optimization of the network parameters, we use Adam~\cite{kingma2014adam} with $\beta _1 = 0.9, \beta _2 = 0.999$,  and the learning rate is initialized to $0.0001$. 
For three model variants, RSTCANet-B,S and L, the learning rate decreases by half each 40k, 
%iterations, 
100k, 
%iterations, 
and 200k iterations, respectively.
Here k equals to $1000$.
The window size is set to 8 by default.
Other parameter settings can be found in Table.~\ref{tab:mv}.

\subsection{Ablation Study and Discussion}\label{subsec:as}

Table~\ref{tab:as1} shows the results of ablation study of RSTCAB. 
We have investigated the effect of Multihead size (MS), short skip connection (SSC), and the number of channel attention (CA) blocks in one RSTCAB. 
 We have selected the RSTCANet trained with 2 RSTCAB, Multihead size 4, three channel attention blocks in one RSTCAB, channel number (C) 64 as the benchmark (see  Table~\ref{tab:as1}).
All models are evaluated for image demosaicing on McM dataset~\cite{zhang2011color} by two metrics, cPSNR and SSIM.

\begin{table}[!ht]
    \centering
    \caption{The ablation study of different components of RSTCAB.}
    \label{tab:as1}
    \resizebox{0.8\linewidth}{!}{
    \begin{tabular}{l|c c c c }
    \toprule
    Case & MS & SSC & CA & cPSNR/SSIM \\\hline
    \hline
    RSTCANet & 4 & \XSolidBrush & 3 & 38.71/0.9897 \\\hline
    RSTCANet-h2 & 2 & \XSolidBrush & 3 & 38.55/0.9892 \\
    \hline
    RSTCANet-SSC & 4 & \Checkmark & 3 & 38.60/0.9896 \\
    \hline
    RSTCANet-CA0 & 4 & \XSolidBrush & 0 & 38.66/0.9897 \\
    RSTCANet-CA1 & 4 & \XSolidBrush & 1 & 38.68/0.9896 \\
    RSTCANet-CA6 & 4 & \XSolidBrush & 6 & 38.65/0.9896 \\
    \bottomrule
    \end{tabular}}
\end{table}

\textbf{Impact of Multihead size.} We have  designed a RSTCANet-h2, with size of Multihead equals to 2.
It can be observed that the model performance can be improved by a bigger attention Multihead size with the same channel number.

\textbf{Impact of short skip connection.} The RSTCANet-SSC is designed to check if adding a short skip connection for every two successive STLs would provide any improvement.   By comparing
%Compared 
RSTCANet and RSTCANet-SSC, one can see that extra skip connection reduces the performance of RSTCAB.

\textbf{The impact of the number of channel attention blocks in one RSTCAB.} Three other variations of RSTCANet are designed to examine the effect of CA blocks. 
There are no CA blocks in RSTCANet-CA0.
Note that the structure of RSTCANet-CA0 is identical to the structure of SwinIR~\cite{liang2021swinir}. In the RSTCAB of RSTCANet-CA1 
there is only one CA block, and the input of this CA block is the input of RSTCAB.
The attention generated by this CA block is multiplied by the features produced by the sixth STL of RSTCAB.
For RSTCANet-CA6, in the RSTCAB, there is one CA block for each STL.

A comparison with RSTCANet-CA0 presented in Table~\ref{tab:as1}, shows that by  exploiting one CA block for six STLs (RSTCANet-CA1) or applying one CA block for each two successive STLs (RSTCANet) can improve the performance of the  RSTCAB.
%A positive effect of channel attention blocks is demonstrated by the obvious performance improvement.
It also demonstrated  that the proposed RSTCANet outperforms another Swin Transformer-based method SwinIR on image demosaicing.
However, when there are six CA blocks in  RSTCAB (RSTCANet-CA6), a performance of RSTCANet becomes worse, which can be explained by
%It should because 
the shifted window partitioning mechanism for two successive STLs~\cite{liu2021swin}.
To make up for the deficiency of cross-window connections in the window-based self-attention module, the authors of \cite{liu2021swin} introduced the shifted window partitioning strategy in two successive Swin transformer blocks.
When the channel attention is learned for each STL, the connections across windows are ignored.
In contrast, by applying the channel attention for every two successive STLs, there is a positive  effect of the shifted window partitioning strategy.

\begin{table*}[!htb]
    \centering
    \caption{The quantitatively comparison with state-of-the-art methods for demosaicing on benchmark datasets. The last column is the size of the model. The best values are in  \textbf{bold}.}
    \label{tab:compare}
    \resizebox{0.65\linewidth}{!}{
    \begin{tabular}{l|c c c c | c}
    \toprule
    \multirow{2}{*}{Method} & McM~\cite{zhang2011color} & Kodak~\cite{kodak1993kodak} & CBSD68~\cite{martin2001database} & Urban100~\cite{huang2015single} & Size\\
    & cPSNR/SSIM & cPSNR/SSIM & cPSNR/SSIM & cPSNR/SSIM & (MB)\\\hline
    \hline
    IRCNN~\cite{zhang2017learning} & 37.84/0.9885 & 40.65/0.9915 & 40.31/0.9924 & 37.03/0.9864 & 18.0\\
    DRUNet~\cite{zhang2021plug} & 39.40/0.9914 & 42.30/0.9944 & 42.33/0.9955 & 39.22/0.9906 & 124.5\\
    RNAN~\cite{zhang2019residual} & 39.66/0.9915 & \textbf{42.92}/\textbf{0.9952} & 42.45/0.9959 & 39.65/0.9923 & 34.3\\\hline
    RSTCANet-B & 38.89/0.9902 & 42.11/0.9948 & 41.74/0.9954 & 38.52/0.9906 & 5.5\\
    RSTCANet-S & 39.58/0.9910 & 42.61/0.9951 & 42.36/0.9958 & 39.69/0.9924 & 16.0\\
    RSTCANet-L & \textbf{39.91}/\textbf{0.9916} & 42.74/\textbf{0.9952} & \textbf{42.47}/\textbf{0.9960} & \textbf{40.07}/\textbf{0.9931} & 32.6\\
    \bottomrule
    \end{tabular}}
     \vspace{-0.1cm}
\end{table*}

\begin{table}[!htb]
    \centering
    \caption{The ablation study of convolutional layers in RSTCANet.
    Image demosaicing on McM dataset.
    $\#$Conv. represent the number of convolution layers. DFE denotes the deep feature extraction module.}
    \label{tab:as2}
    \resizebox{\linewidth}{!}{
    \begin{tabular}{l|c c c c }
    \toprule
    \multirow{2}*{Case} & \multicolumn{2}{c}{$\#$Conv. in} & \multirow{2}*{cPSNR/SSIM} & $\#$param. \\
    & RSTCAB & DFE & &(MB)\\\hline
    \hline
    RSTCANet-B & 1 & 1 & 38.89/0.9902 & 5.5 \\\hline
    RSTCANet-1 & 1 & 2 & 38.88/0.9899 & 5.7 \\\hline
    RSTCANet-2 & 2 & 1 & 38.82/0.9898 & 5.9 \\\hline
    RSTCANet-3 & 0 & 1 & 38.52/0.9894 & 5.1 \\
    \bottomrule
    \end{tabular}}
    \vspace{-0.1cm}
\end{table}

\begin{table}[!htb]
    \centering
    \caption{The ablation study of basic block of demosaicing network.
    Image demosaicing on McM dataset.}
    \label{tab:as3}
    \resizebox{\linewidth}{!}{
    \begin{tabular}{l|c c c c }
    \toprule
    Case & Basic Block & cPSNR/SSIM & $\#$param.(MB)\\\hline
    \hline
    RSTCANet-B & RSTCAB & 38.89/0.9902 & 5.5 \\\hline
    SwinIR$^*$ & RSTB~\cite{liang2021swinir} & 38.76/0.9898 & 5.5 \\\hline
    RCAN$^*$ & RCAB~\cite{zhang2018image} & 38.86/0.9899 & 7.7 \\\hline
    SCNet$^*$ & SC block~\cite{zhang2022practical} & 39.00/0.9901 & 8.5\\
    \bottomrule
    \end{tabular}}
    \vspace{-0.1cm}
\end{table}

\textbf{Impact of Convolutional layers.} We have also 
%try to add more 
added extra convolution layers in the RSTCANet,  
%We choose two positions to add more convolution layers: 
at the end of RSTCAB and at the end of deep feature extraction module. 
We have trained these two variations of RSTCANet, and denoted them as RSTCANet-1 and RSTCANet-2, respectively. Another variation, RSTCANet-3 without convolutions in the RSTCAB is trained as well. All models in this experiment are
%is 
trained with channel features 72 and Multihead size 6 as RSTCANet-B. 
From the results  presented in Table~\ref{tab:as2}, we unexpectedly find that utilization of more convolution layers in RSTCANet does not guarantee the performance improvement, but 
%must led 
 leads to an increase of the model size.% bigger.
However, the experimental result demonstrated   (RSTCANet-3) that the convolutional layer at the end of RSTCAB is necessary. 
It should be the case because the STL layers are more relevant for recognition rather than reconstruction~\cite{he2021masked}.

\textbf{U-Net or not.}
The U-Net architecture is very popular in image restoration networks.
One may expect that a combination of Swin Transformer and U-Net  can further improve the performance of  model~\cite{cao2021swin,fan2022sunet,zhang2022practical}.
We have  trained a model with U-Net structure, denoted as RSTCAUNet.
Except for the U-Net structure, all other settings are identical, including the number of RSTCAB.
The RSTCAUNet model size is  218.2 MB and  for image demosaicing on McM dataset it gets 36.92 dB.
In contrast, the compared RSTCANet model has only 16.8 MB but better demosaicing performance (38.28 dB).
This comparison shows  that the U-Net structure is not suitable for RSTCANet for demosaicing.

\textbf{Impact of the  Basic Block.} 
 A comparison between RSTCANet and RSTCANet-CA0 in Table~\ref{tab:as1} shows that RSTCAB performs better than RSTB~\cite{liang2021swinir} on demosaicing.
Besides this, we train another RSTB-based model with a bigger channel number (72), denoted as SwinIR$^*$.
We also compare our proposed RSTCAB with other two related basic blocks, RCAB~\cite{zhang2018image} and SC block~\cite{zhang2022practical}; we marked these models as RCAN$^*$ and SCNet$^*$, respectively. 
The number of RCAB in RCAN$^*$ is 24, 
and the number of SC blocks in SCNet$^*$ is 2 and each SC block has 6 Swin Transformer blocks.
For fair comparison, all other training settings of SwinIR$^*$, RCAN$^*$ and SCNet$^*$ are same with the proposed RSTCANet-B, including the structure of shallow feature extraction module and the image reconstruction module.
From Table~\ref{tab:as3}, one can see that combining the CA blocks whose parameters are shared with the Swin Transformer blocks can improve the demosaicing performance of model without additional storage cost.
The numbers of parameters of RSTCANet-B and SwinIR$^*$ are same,
but the demosaicing performance is improved by 0.13 dB.
In addition, the model size of RCAN$^*$ is $1.4$ times of RSTCANet-B.
While the number of parameters is more, the demosaicing performance of RCAN$^*$ on McM set is slightly worse than RSTCANet-B.
This shows that the proposed RSTCAB combines advantages of RCAB and RSTB. 
For SCNet$^*$, it achieves a slight  improvement (0.11 dB) at the cost of $1.4$ times parameters
of RSTCANet-B.
This demonstrates that RSTCAB has a better trade-off between the demosaicing performance and model size than SC block.

\subsection{Results on Image Demosaicing}

\textbf{Quantitative comparison.} Table~\ref{tab:compare} shows the quantitative comparisons between RSTCANet and the state-of-the-art methods: DRUNet~\cite{zhang2021plug}, IRCNN~\cite{zhang2017learning} and RNAN~\cite{zhang2019residual}. We test different methods on four benchmark datasets, McM~\cite{zhang2011color}, Kodak~\cite{kodak1993kodak}, CBSD68~\cite{martin2001database} and Urban100~\cite{huang2015single}. The color PSNR and SSIM values are evaluated on resulting images. 

As one can see, compared with these state-of-the-art  demosaicing methods, our RSTCANet can get comparable performance with smaller size. 
The RSTCANet-B performs better than IRCNN at least 1 dB with 0.3$\times$ model size.
RSTCANet-S outperforms DRUNet by 0.18 dB on McM, 0.31 dB on Kodak and 0.47 dB on Urban100 with only $0.13\times$ of its size. 
On CBSD68, our RSTCANet-S also performs slightly better than DRUNet.
Compared with the SOTA method RNAN, our RSTCANet-S gets the comparable performance with only half size parameters. 
On Urban100, the RSTCANet-S even achieves a slightly better performance (0.04 dB) than RNAN.

The large model variant, RSTCANet-L, is a lighter (1.7 MB)  than RNAN,
but has better performance than RNAN on three datasets. 
Especially on Urban100, our RSTCANet-L performs better than RNAN by 0.42 dB.

In addition, by increasing the channel number C and the number of RSTCAB blocks, RSTCANet-S and RSTCANet-L improve the performance by at least 0.5 dB and 0.63 dB compared with RSTCANet-B.

% \begin{figure*}[!ht]
% \centering
% \sbox{\measurebox}{
%   \begin{minipage}[b]{.59\textwidth}
%   \subfloat{\includegraphics[width=\textwidth]{RSTCANet/fig/ri/kodim19_GT_draw.png}}
%   \caption*{Kodak:kodim19.png}
%   \end{minipage}}
% \usebox{\measurebox}\enspace
% %
% \begin{minipage}[b]{.18\textwidth}
% \centering
% \subfloat{\includegraphics[width=\textwidth]{RSTCANet/fig/ri/kodim19_GT.png}}
% \caption*{Ground truth}
% %\vfill

% \subfloat{\includegraphics[width=\textwidth]{RSTCANet/fig/ri/kodim19_ircnn_color.png}}
% \caption*{IRCNN}

% \subfloat{\includegraphics[width=\textwidth]{RSTCANet/fig/ri/kodim19_drunet_color.png}}
% \caption*{DRUNet}

% \subfloat{\includegraphics[width=\textwidth]{RSTCANet/fig/ri/kodim19_RNAN.png}}
% \caption*{RNAN}
% \end{minipage}
% %
% \begin{minipage}[b]{.18\textwidth}
% \centering
% \subfloat{\includegraphics[width=\textwidth]{RSTCANet/fig/ri/kodim19.png}}
% \caption*{Mosaiced}
% %\vfill

% \subfloat{\includegraphics[width=\textwidth]{RSTCANet/fig/ri/kodim19_RSTCANetB.png}}
% \caption*{RSTCANet-B}

% \subfloat{\includegraphics[width=\textwidth]{RSTCANet/fig/ri/kodim19_RSTCANetS.png}}
% \caption*{RSTCANet-S}

% \subfloat{\includegraphics[width=\textwidth]{RSTCANet/fig/ri/kodim19_RSTCANetL.png}}
% \caption*{RSTCANet-L}
% \end{minipage}
% \caption{Visual results comparison of different demosaicing methods on image \textit{kodim19} from Kodak dataset.}
% \label{fig:dm_kodak_compare}
% \end{figure*}

\textbf{Visual comparison.} Fig.~\ref{fig:dm_urban_compare} illustrates the visual comparisons between our proposed RSTCANet and the state-of-the-art demosaicing methods. 
In Fig.~\ref{fig:dm_urban_compare}, one can observe that the proposed RSTCANet-L can generate less color artifacts than other methods. 
For high frequency regions, the color artifacts exist even in 
%the SOTA method 
RNAN resulting images.
In contrast, our method can reconstruct the color image with less color artifacts.

\section{Conclusion}
In this paper, we propose a Swin Transformer-based image demosaicing model RSTCANet,  %.
based on the
%For deep feature extraction, we propose the 
residual Swin Transformer Channel Attention blocks (RSTCAB), 
which takes advantage of both Swin Transformer 
%blocks 
and Channel Attention blocks.
Experimental results show that RSTCAB surpass other Swin Transformer-based blocks on image demosaicing.
The quantitative and qualitative results also demonstrate that 
RSTCANet achieves state-of-the-art performance on image demosaicing, generating much less color artifacts in the resulting images.
In the future, we plan to extend the RSTCANet to other image restoration tasks, such as image denoising and super-resolution.
\clearpage

%\section*{Acknowledgment}

\bibliographystyle{IEEEtran}
\bibliography{ref}

\begin{thebibliography}{10}
\providecommand{\url}[1]{#1}
\csname url@samestyle\endcsname
\providecommand{\newblock}{\relax}
\providecommand{\bibinfo}[2]{#2}
\providecommand{\BIBentrySTDinterwordspacing}{\spaceskip=0pt\relax}
\providecommand{\BIBentryALTinterwordstretchfactor}{4}
\providecommand{\BIBentryALTinterwordspacing}{\spaceskip=\fontdimen2\font plus
\BIBentryALTinterwordstretchfactor\fontdimen3\font minus
  \fontdimen4\font\relax}
\providecommand{\BIBforeignlanguage}[2]{{%
\expandafter\ifx\csname l@#1\endcsname\relax
\typeout{** WARNING: IEEEtran.bst: No hyphenation pattern has been}%
\typeout{** loaded for the language `#1'. Using the pattern for}%
\typeout{** the default language instead.}%
\else
\language=\csname l@#1\endcsname
\fi
#2}}
\providecommand{\BIBdecl}{\relax}
\BIBdecl

\bibitem{hirakawa2005adaptive}
K.~Hirakawa and T.~W. Parks, ``Adaptive homogeneity-directed demosaicing
  algorithm,'' \emph{IEEE Transactions on Image Processing}, vol.~14, no.~3,
  pp. 360--369, 2005.

\bibitem{malvar2004high}
H.~S. Malvar, L.-w. He, and R.~Cutler, ``High-quality linear interpolation for
  demosaicing of bayer-patterned color images,'' in \emph{2004 IEEE
  International Conference on Acoustics, Speech, and Signal Processing},
  vol.~3.\hskip 1em plus 0.5em minus 0.4em\relax IEEE, 2004, pp. iii--485.

\bibitem{su2006highly}
C.-Y. Su, ``Highly effective iterative demosaicing using weighted-edge and
  color-difference interpolations,'' \emph{IEEE Transactions on Consumer
  Electronics}, vol.~52, no.~2, pp. 639--645, 2006.

\bibitem{zhang2005color}
L.~Zhang and X.~Wu, ``Color demosaicking via directional linear minimum mean
  square-error estimation,'' \emph{IEEE Transactions on Image Processing},
  vol.~14, no.~12, pp. 2167--2178, 2005.

\bibitem{he2012self}
F.-L. He, Y.-C.~F. Wang, and K.-L. Hua, ``Self-learning approach to color
  demosaicking via support vector regression,'' in \emph{2012 19th IEEE
  International Conference on Image Processing}.\hskip 1em plus 0.5em minus
  0.4em\relax IEEE, 2012, pp. 2765--2768.

\bibitem{sun2012separable}
J.~Sun and M.~F. Tappen, ``Separable markov random field model and its
  applications in low level vision,'' \emph{IEEE transactions on image
  processing}, vol.~22, no.~1, pp. 402--407, 2012.

\bibitem{go2000interpolation}
J.~Go, K.~Sohn, and C.~Lee, ``Interpolation using neural networks for digital
  still cameras,'' \emph{IEEE Transactions on Consumer Electronics}, vol.~46,
  no.~3, pp. 610--616, 2000.

\bibitem{kapah2000demosaicking}
O.~Kapah and H.~Z. Hel-Or, ``Demosaicking using artificial neural networks,''
  in \emph{Applications of Artificial Neural Networks in Image Processing V},
  vol. 3962.\hskip 1em plus 0.5em minus 0.4em\relax International Society for
  Optics and Photonics, 2000, pp. 112--120.

\bibitem{syu2018learning}
N.-S. Syu, Y.-S. Chen, and Y.-Y. Chuang, ``Learning deep convolutional networks
  for demosaicing,'' \emph{arXiv preprint arXiv:1802.03769}, 2018.

\bibitem{gharbi2016deep}
M.~Gharbi, G.~Chaurasia, S.~Paris, and F.~Durand, ``Deep joint demosaicking and
  denoising,'' \emph{ACM Transactions on Graphics (ToG)}, vol.~35, no.~6, pp.
  1--12, 2016.

\bibitem{zhang2021plug}
K.~Zhang, Y.~Li, W.~Zuo, L.~Zhang, L.~Van~Gool, and R.~Timofte, ``Plug-and-play
  image restoration with deep denoiser prior,'' \emph{IEEE Transactions on
  Pattern Analysis and Machine Intelligence}, 2021.

\bibitem{zhang2017learning}
K.~Zhang, W.~Zuo, S.~Gu, and L.~Zhang, ``Learning deep cnn denoiser prior for
  image restoration,'' in \emph{Proceedings of the IEEE conference on computer
  vision and pattern recognition}, 2017, pp. 3929--3938.

\bibitem{zhang2019residual}
Y.~Zhang, K.~Li, K.~Li, B.~Zhong, and Y.~Fu, ``Residual non-local attention
  networks for image restoration,'' \emph{arXiv preprint arXiv:1903.10082},
  2019.

\bibitem{liu2021swin}
Z.~Liu, Y.~Lin, Y.~Cao, H.~Hu, Y.~Wei, Z.~Zhang, S.~Lin, and B.~Guo, ``Swin
  transformer: Hierarchical vision transformer using shifted windows,''
  \emph{arXiv preprint arXiv:2103.14030}, 2021.

\bibitem{cao2021swin}
H.~Cao, Y.~Wang, J.~Chen, D.~Jiang, X.~Zhang, Q.~Tian, and M.~Wang,
  ``Swin-unet: Unet-like pure transformer for medical image segmentation,''
  \emph{arXiv preprint arXiv:2105.05537}, 2021.

\bibitem{liang2021swinir}
J.~Liang, J.~Cao, G.~Sun, K.~Zhang, L.~Van~Gool, and R.~Timofte, ``Swinir:
  Image restoration using swin transformer,'' in \emph{Proceedings of the
  IEEE/CVF International Conference on Computer Vision}, 2021, pp. 1833--1844.

\bibitem{zhang2018image}
Y.~Zhang, K.~Li, K.~Li, L.~Wang, B.~Zhong, and Y.~Fu, ``Image super-resolution
  using very deep residual channel attention networks,'' in \emph{Proceedings
  of the European conference on computer vision (ECCV)}, 2018, pp. 286--301.

\bibitem{zhang2022practical}
K.~Zhang, Y.~Li, J.~Liang, J.~Cao, Y.~Zhang, H.~Tang, R.~Timofte, and
  L.~Van~Gool, ``Practical blind denoising via swin-conv-unet and data
  synthesis,'' \emph{arXiv preprint arXiv:2203.13278}, 2022.

\bibitem{zhang2020residual}
Y.~Zhang, Y.~Tian, Y.~Kong, B.~Zhong, and Y.~Fu, ``Residual dense network for
  image restoration,'' \emph{IEEE Transactions on Pattern Analysis and Machine
  Intelligence}, vol.~43, no.~7, pp. 2480--2495, 2020.

\bibitem{Agustsson_2017_CVPR_Workshops}
E.~Agustsson and R.~Timofte, ``Ntire 2017 challenge on single image
  super-resolution: Dataset and study,'' in \emph{The IEEE Conference on
  Computer Vision and Pattern Recognition (CVPR) Workshops}, July 2017.

\bibitem{kingma2014adam}
D.~P. Kingma and J.~Ba, ``Adam: A method for stochastic optimization,''
  \emph{arXiv preprint arXiv:1412.6980}, 2014.

\bibitem{zhang2011color}
L.~Zhang, X.~Wu, A.~Buades, and X.~Li, ``Color demosaicking by local
  directional interpolation and nonlocal adaptive thresholding,'' \emph{Journal
  of Electronic imaging}, vol.~20, no.~2, p. 023016, 2011.

\bibitem{kodak1993kodak}
E.~Kodak, ``Kodak lossless true color image suite (photocd pcd0992),''
  \emph{URL http://r0k. us/graphics/kodak}, vol.~6, 1993.

\bibitem{martin2001database}
D.~Martin, C.~Fowlkes, D.~Tal, and J.~Malik, ``A database of human segmented
  natural images and its application to evaluating segmentation algorithms and
  measuring ecological statistics,'' in \emph{Proceedings Eighth IEEE
  International Conference on Computer Vision. ICCV 2001}, vol.~2.\hskip 1em
  plus 0.5em minus 0.4em\relax IEEE, 2001, pp. 416--423.

\bibitem{huang2015single}
J.-B. Huang, A.~Singh, and N.~Ahuja, ``Single image super-resolution from
  transformed self-exemplars,'' in \emph{Proceedings of the IEEE conference on
  computer vision and pattern recognition}, 2015, pp. 5197--5206.

\bibitem{he2021masked}
K.~He, X.~Chen, S.~Xie, Y.~Li, P.~Doll{\'a}r, and R.~Girshick, ``Masked
  autoencoders are scalable vision learners,'' \emph{arXiv preprint
  arXiv:2111.06377}, 2021.

\bibitem{fan2022sunet}
C.-M. Fan, T.-J. Liu, and K.-H. Liu, ``Sunet: Swin transformer unet for image
  denoising,'' \emph{arXiv preprint arXiv:2202.14009}, 2022.

\end{thebibliography}
\end{document}